%% file: main.tex
\documentclass{article}

\PassOptionsToPackage{numbers}{natbib}

\usepackage[final]{neurips_2019}

\usepackage[utf8]{inputenc} 
\usepackage[T1]{fontenc}    
\usepackage{hyperref}       
\usepackage{url}            
\usepackage{booktabs}       
\usepackage{amsfonts}       
\usepackage{nicefrac}       
\usepackage{microtype}      

\usepackage[noabbrev,capitalize]{cleveref}
\usepackage{caption}
\usepackage{subcaption}

\usepackage{amsmath}
\usepackage{booktabs}
\usepackage{times}
\usepackage{multicol}
\usepackage{graphicx}
\usepackage{geometry}
\usepackage{array, caption, floatrow, tabularx, makecell, booktabs}%
\setcellgapes{3pt}
\usepackage{mlmacros}
\usepackage[T1]{fontenc}
\usepackage[acronym,smallcaps,nowarn,section,nogroupskip,nonumberlist]{glossaries}

\glsdisablehyper 
\input{glossary}
\renewcommand{\citep}[1]{(\citet{#1})}
\variables{a,g,t,x,y,z}

\newcommand\blfootnote[1]{%
  \begingroup
  \renewcommand\thefootnote{}\footnote{#1}%
  \addtocounter{footnote}{-1}%
  \endgroup
}

\title{End-to-End Recurrent Multi-Object Tracking and Trajectory Prediction with Relational Reasoning}

\author{Fabian B. Fuchs,\, Adam R. Kosiorek,\, Li Sun,\, Oiwi Parker Jones,\,Ingmar Posner\\
Applied AI Lab,  University of Oxford \\
\texttt{\{fabian, adamk, kevin, oiwi, ingmar\}@robots.ox.ac.uk} \\
}

\begin{document}

\maketitle

\begin{abstract}

The majority of contemporary object-tracking approaches do not model interactions between objects.
This contrasts with the fact that objects' paths are not independent: a cyclist might abruptly deviate from a previously planned trajectory in order to avoid colliding with a car.
Building upon \textsc{hart}, a neural class-agnostic single-object tracker, we introduce a multi-object tracking method (\textsc{mohart}) capable of \emph{relational reasoning}.
Importantly, the entire system, including the understanding of interactions and relations between objects, is class-agnostic and learned simultaneously in an end-to-end fashion.
We explore a number of relational reasoning architectures and show that permutation-invariant models outperform non-permutation-invariant alternatives. We also find that architectures using a single permutation invariant operation like DeepSets, despite, in theory, being universal function approximators, are nonetheless outperformed by a more complex architecture based on multi-headed attention. The latter better accounts for complex physical interactions in a challenging toy experiment.
Further, we find that modelling interactions leads to consistent performance gains in tracking as well as future trajectory prediction on three real-world datasets (MOTChallenge, UA-DETRAC, and Stanford Drone dataset), particularly in the presence of ego-motion, occlusions, crowded scenes, and faulty sensor inputs.

\end{abstract}

\blfootnote{This is an extended version of the paper "Permutation Invariance and Relational Reasoning in Multi-Object Tracking" presented at the \textit{Sets and Partitions Workshop} at the \textit{33rd Conference on Neural Information Processing Systems, Vancouver 2019}}
\input{1_introduction}

\input{1b_related_work}

\input{2_model}

\input{3_validation}

\input{4_experiments}
\input{5_discussion}
\section*{Acknowledgements}
We thank Stefan Saftescu for his contributions, particularly for integrating the Stanford Drone Dataset. We thank Jonathon Luiten for providing baselines on the MOTChallenge dataset and Adam Golinski as well as Stefan Saftescu for proof-reading. This research was funded by the EPSRC AIMS Centre for Doctoral Training at Oxford University and an EPSRC Programme Grant (EP/M019918/1).
We acknowledge use of Hartree Centre resources in this work. The STFC Hartree Centre is a research collaboratory in association with IBM providing High Performance Computing platforms funded by the UK's investment in e-Infrastructure. The Centre aims to develop and demonstrate next generation software, optimised to take advantage of the move towards exa-scale computing.
{
	\small
	\bibliographystyle{plainnat}
	\newpage\bibliography{library.bib}  
}
\newpage\clearpage
\appendix
\input{6_experimental_details.tex}

\input{6_architecture_details.tex}
\input{6_first_toy_experiment.tex}
\input{6_exp_prediction.tex}
\input{6_blackout.tex}

\end{document}

%% file: glossary.tex
\newacronym{VAE}{vae}{variational auto-encoder}
\newacronym{SSM}{ssm}{state-space model}
\newacronym{SGD}{sgd}{stochastic gradient descent}
\newacronym{ELBO}{elbo}{evidence lower bound}
\newacronym{KL}{kl}{Kullback-Leibler}
\newacronym{LSTM}{lstm}{long short-term memory}
\newacronym{CNN}{cnn}{convolutional neural network}

\newacronym{ADAM}{adam}{adam}
\glsunset{ADAM}
\newacronym{RMSprop}{rmsprop}{RMSprop}
\glsunset{RMSprop}

\newacronym{GAN}{gan}{generative adversarial network}
\newacronym{GRU}{gru}{gated recurrent unit}
\newacronym{MLP}{mlp}{multilayer perceptron}
\newacronym{MC}{mc}{Monte Carlo}
\newacronym[firstplural=recurrent neural networks, plural=RNNs]{RNN}{rnn}{recurrent neural network}

\newacronym{MNIST}{mnist}{mnist}
\glsunset{MNIST}

\newacronym[firstplural=degrees of freedom, plural=DoFs]{DOF}{dof}{degree of freedom}

\newacronym{GENESIS}{genesis}{\textsc{gene}rative \textsc{s}cene \textsc{i}nference and \textsc{s}ampling}

\newacronym{GMM}{gmm}{gaussian mixture model}
\newacronym{SBP}{sbp}{stick-breaking process}

\newacronym{HART}{hart}{hierarchical attentive recurrent tracking}
\newacronym{MOHART}{mohart}{multi-object hierarchical attentive recurrent tracking}

\newacronym{SOT}{sot}{single-object tracking}
\newacronym{MOT}{mot}{multi-object tracking}
\newacronym{VOT}{vot}{visual object tracking}

\newacronym{SOTA}{sota}{state-of-the-art}
\newacronym{RPN}{rpn}{region proposal network}
\newacronym{IOU}{iou}{intersection-over-union}

\newacronym{RCNN}{rcnn}{rcnn}
\glsunset{RCNN}
\newacronym{SIAMRCNN}{siamrcnn}{siamrcnn}
\glsunset{SIAMRCNN}
\newacronym{SIAMRPN}{siamrpn++}{siamrpn++}
\glsunset{SIAMRPN}
\newacronym{DIMP}{dimp}{dimp}
\glsunset{DIMP}
\newacronym{ECO}{eco}{eco}
\glsunset{ECO}

%% file: 1_introduction.tex
\section{Introduction}
\label{sec:intro}


It is imperative that any autonomous agent acting in the real world be capable of accounting for a variety of present objects and for interactions between these objects.
This motivates the need for tracking algorithms which are \emph{class-agnostic} and able to model the dynamics of multiple objects---properties not yet supported by current state-of-the-art visual object trackers.
These often use detectors or region proposal networks such as Faster-\gls{RCNN} \cite{RCNN, FasterRCNN}.
Algorithms from this family can achieve high accuracy, provided sufficient labelled data to train the object detector, and given that all encountered objects can be associated with known classes, but fail when faced with objects from unseen categories.

\Gls{HART} \cite{Kosiorek17} is a recently-proposed, alternative method for single-object tracking (\textsc{sot}), which can track arbitrary objects indicated by the user.
As is common in \gls{VOT}, \gls{HART} is provided with a bounding box in the first frame. In the following frames,
%
\gls{HART} efficiently processes just the relevant part of an image using spatial attention; it also integrates object detection, feature extraction, and motion modelling into one end-to-end network.
Contrary to most methods,
which process video frames one at a time,
end-to-end learning in \gls{HART} allows for discovering complex visual and spatio-temporal patterns in videos, which is conducive to inferring what an object is and how it moves. It is also class-agnostic as it does not rely on a pre-trained detector.

\begin{figure}[t!]
	\centering
	\includegraphics[width=\linewidth]{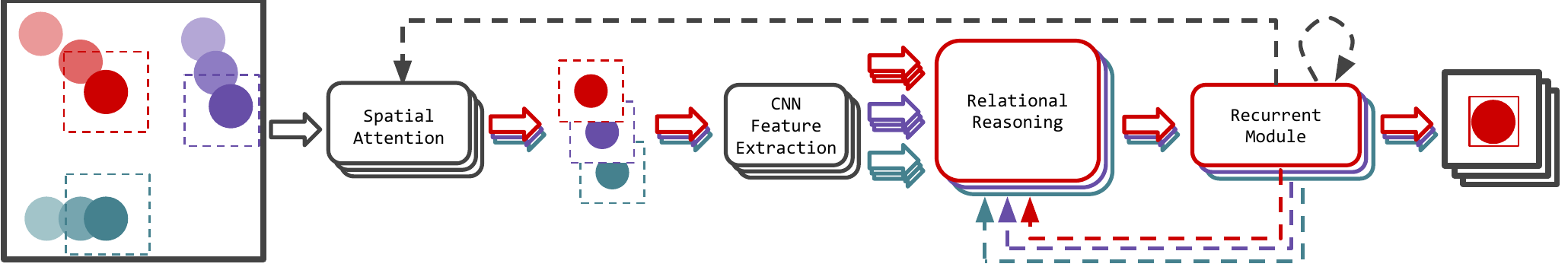}
	\vspace{-4mm}
	\caption{
		\Gls{MOHART}. A glimpse is extracted for each object using a (fully differentiable) spatial attention mechanism. These glimpses are further processed with a CNN and fed into a relational reasoning module. A recurrent module, which iterates over time steps, allows for capturing of complex motion patterns. It also outputs spatial attention parameters and a feature vector per object for the relational reasoning module. Dashed lines indicate temporal connections (from time step $t$ to $t+1$). The entire pipeline operates in parallel for the different objects, only the relational reasoning module allows for exchange of information between tracking states of each object. \gls{MOHART} is an extension of \acrshort{HART} (a single-object tracker), which features the same pipeline without the relational reasoning module.
	}
	\label{fig:teaser}
\end{figure}

In the original formulation, \gls{HART} is limited to the single-object modality---as are other existing end-to-end trackers \citep{Kahou15,Danesh19,Gordon2018}.
In this work, we present \gls{MOHART}, a class-agnostic tracker with 
relational reasoning capabilities.
\Gls{MOHART} infers the latent state of every tracked object in parallel, informing per-object states about other tracked objects using self-attention \citep{Vaswani17,Lee2019settransformer}.
This helps to avoid performance loss under self-occlusions of tracked objects or strong camera motion.
Moreover, since the model is trained end-to-end, it is able to learn how to manage faulty or missing sensor inputs. See \Cref{fig:teaser} for a high-level illustration of \gls{MOHART}. The relational-reasoning module receives a list of feature vectors, one per object, as input. This part of the problem is permutation invariant, as the list-order of object representations carries no meaning for the task at hand. 

In order to track objects, \gls{MOHART} estimates their states, which can be naturally used to predict future trajectories over short temporal horizons, which is especially useful for planning in the context of autonomous agents.
\gls{MOHART} can be trained simultaneously for object tracking and trajectory prediction, thereby increasing statistical efficiency of learning. 
In contrast to prior art, where these two tasks are usually addressed as separate problems with unrelated solutions, our work shows trajectory prediction and object tracking are best addressed jointly.

%% file: 1b_related_work.tex
\section{Related Work}
\label{sec:related}

\paragraph{Visual Object Tracking \textsc{vot}} In \gls{VOT}, a ground-truth bounding box is provided to the algorithm in the first frame and the tracker is evaluated on all future frames. The performance is typically measured as \gls{IOU} averaged across all frames in which the object is present. On many \gls{VOT} datasets, \gls{SIAMRCNN} \cite{SiamRCNN} is currently \gls{SOTA}. In each frame, it employs a pre-trained \gls{RPN} providing candidate bounding boxes, which are then compared to the bounding box in the first frame. Previous \gls{SOTA} models include \gls{SIAMRPN} \cite{SiamRPNpp}
\gls{ECO} \cite{ECO}, and \gls{DIMP} \cite{DIMP}. 
These models are highly engineered achieving excellent results: They often use \gls{RPN}s pre-trained on large amounts of data and fine-tune their model online on the target dataset. 
All of these models are tracking one object at a time (although \gls{SIAMRCNN} does track distractor objects) and therefore do not perform relational reasoning. They also do not have internal motion models of the objects.

\vspace{-1em} 
\paragraph{End-to-End \gls{VOT}} A newly established and much less explored stream of work approaches region proposal, feature extraction and tracking in an end-to-end fashion with gradients propagated through all parts of the model and across the time axis. This allows for complex motion models as well as efficiency (learning where to look). A key difficulty here is that extracting an image crop (according to bounding-boxes provided by a detector), is non-differentiable and results in high-variance gradient estimators.
\citet{Kahou15} propose an end-to-end tracker with soft spatial-attention using a 2D grid of Gaussians instead of a hard bounding-box. \Gls{HART} draws inspiration from this idea, employs an additional attention mechanism, and shows promising performance on the real-world KITTI dataset \cite{Kosiorek17}.
\Gls{HART} forms the foundation of this work. It has also been extended to incorporate depth information from \textsc{rgbd} cameras \cite{Danesh19}. \citet{Gordon2018} propose an approach in which the crop corresponds to the scaled-up previous bounding-box. This simplifies the approach but does not allow the model to learn where to look---i.e. no gradient is backpropagated through crop coordinates.
To the best of our knowledge, there are no successful implementations of any such end-to-end approaches for multi-object tracking from vision beyond generative modelling works \cite{Kosiorek2018sqair,Steenkiste2018,Jiang2020scalor} which work only on relatively simple datasets, and \citet{Frossard2018}, which relies on costly \textsc{lidar} sensors.
On real-world data, the only end-to-end approaches correspond to applying multiple single-object trackers in parallel---a method which does not leverage the potential of scene context or inter-object interactions. 
 
\vspace{-1em} 
\paragraph{Tracking-by-Detection} Here, traditionally, objects are first detected in each frame independently. A tracking algorithm then links the detections from different frames to propose a coherent trajectory \cite{Zhang2008,Milan2014,bae2017confidence,keuper2018motion}. 
Often, tracking-by-detection benchmark suites provide external detections for both training and test sets, turning it into a data association task.
Recently, some approaches started reusing the same networks for detecting objects and generating re-identification features \cite{Zhou2020objectsaspoints,Zhang2020asb}.
\Gls{MOHART} currently does not use external detections (provided or from a private detector) and hence cannot be quantitatively compared to these approaches. However, incorporating external detections via the proposed attention framework is a promising future direction of research.


\vspace{-1em} 
\paragraph{Pedestrian trajectory prediction} 
We draw inspiration from this stream of work for developing a relational reasoning module. Social-\textsc{lstm} \cite{social-lstm} employs a \gls{LSTM} to predict pedestrian trajectories and uses max-pooling to model global social context. Attention mechanisms have also been employed to query the most relevant information, such as neighbouring pedestrians, in a learnable fashion  \cite{su2016crowd, fernando2018soft, sadeghian2019sophie}.
Our work stands apart from this prior art by not relying on ground truth tracklets. It addresses the more challenging task of working directly with visual input, performing tracking, modelling interactions, and, depending on the application scenario, simultaneously predicting future motions.

%% file: 2_model.tex
\section{Recurrent Multi-Object Tracking with Self-Attention}
\label{sec:method}

\begin{figure}
	\centering
	\includegraphics[width=\linewidth]{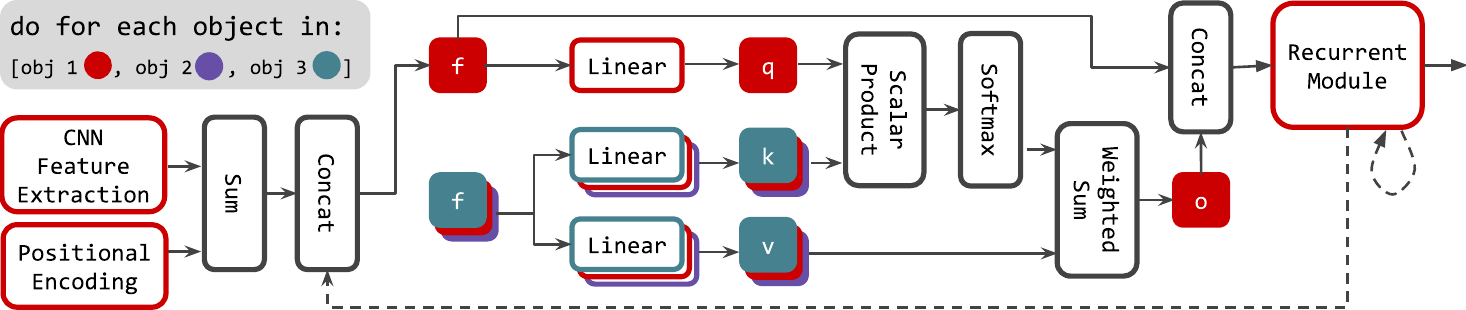}
	\caption{
		The relational reasoning module in \Gls{MOHART} based on multi-headed self-attention. Here, we show the computation of the interaction of the red object with all other objects. Object representations $f_{t,m}$ are computed using visual features, positional encoding and the hidden state from the recurrent module. These are linearly projected onto keys (k), queries (q), and values (v) to compute a weighted sum of interactions between objects, yielding an interaction vector $o_{t,m}$. Subscripts $t, m$ are dropped from all variables for clarity of presentation, so is the splitting into multiple heads.
	}
	\label{fig:sab}
\end{figure}

This section describes the model architecture in \Cref{fig:teaser}. We start by describing the \glsreset{HART}\gls{HART} algorithm \citep{Kosiorek17}, and then follow with an extension of \gls{HART} to tracking multiple objects, where multiple instances of \gls{HART} communicate with each other using multi-headed attention to facilitate relational reasoning. We also explain how this method can be extended to trajectory prediction instead of just tracking. 

\subsection{Hierarchical Attentive Recurrent Tracking (\textsc{hart})}
\textsc{Hart} is an attention-based recurrent algorithm, which can efficiently track single objects in a video.
It uses a spatial attention mechanism to extract a \textit{glimpse} $\bg_t$, which corresponds to a small crop of the image $\bxt$ at time-step $t$, containing the object of interest.
This allows it to dispense with the processing of the whole image and can significantly decrease the amount of computation required.
\Gls{HART} uses a \gls{CNN} to convert the glimpse $\bg_t$ into features $\mathbf{f}_t$, which then update the hidden state $\mathbf{h}_t$ of a \gls{LSTM} core.
The hidden state is used to estimate the current bounding-box $\mathbf{b}_t$, spatial attention parameters for the next time-step $\ba_{t+1}$, as well as object appearance.
Importantly, the recurrent core can learn to predict complicated motion conditioned on the past history of the tracked object, which leads to relatively small attention glimpses---contrary to \gls{CNN}-based approaches \cite{Held2016goturn,Valmadre2017corr}, \gls{HART} does not need to analyse large regions-of-interest to search for tracked objects.
In the original paper, \textsc{hart} processes the glimpse with an additional ventral and dorsal stream.
Early experiments have shown that this does not improve performance on the MOTChallenge dataset, presumably due to the oftentimes small objects and overall small amount of training data. 
Further details are provided in \Cref{sec:architecture_details}.

	The algorithm is initialised with a bounding-box\footnote{We can use either a ground-truth bounding-box or one provided by an external detector; the only requirement is that it contains the object of interest.} $\mathbf{b}_1$ for the first time-step, and operates on a sequence of raw images $\bx_{1:T}$.
	For time-steps $t\geq2$, it recursively outputs bounding-box estimates for the current time-step and predicted attention parameters for the next time-step. The performance is measured as \gls{IOU} averaged over all time steps in which an object is present, excluding the first time step.

\gls{HART} is limited to tracking one object at a time.
While it can  be deployed on several objects in parallel, different \gls{HART} instances have no means of communication.
This results in performance loss, as it is more difficult to identify 
occlusions, ego-motion and object interactions.
Below, we propose an extension of \gls{HART} which remedies these shortcomings.

\subsection{Multi-Object Hierarchical Attentive Recurrent Tracking}
Multi-object support in \gls{HART} requires the following modifications.
Firstly, in order to handle a dynamically changing number of objects, we apply \gls{HART} to multiple objects in parallel, where all parameters between \gls{HART} instances are shared. 
We refer to each \gls{HART} instance as a \textit{tracker}.
Secondly, we introduce a presence variable $p_{t,m}$ for object $m$.
It is used to mark whether an object should interact with other objects, as well as to mask the loss function (described in \citet{Kosiorek17}) for the given object when it is not present.
In this setup, parallel trackers cannot exchange information and are conceptually still single-object trackers, which we use as a baseline, referred to as \gls{HART} (despite it being an extension of the original algorithm).
Finally, to enable communication between trackers, we augment \gls{HART} with an additional step between feature extraction and the \gls{LSTM}.

For each object, a glimpse is extracted and processed by a \gls{CNN} (see \Cref{fig:teaser}). Furthermore, spatial attention parameters are linearly projected on a vector of the same size and added to this representation, acting as a positional encoding. This is then concatenated with the hidden state of the recurrent module of the respective object (see \Cref{fig:sab}). Let $\mathbf{f}_{t, m}$ denote the resulting feature vector corresponding to the m$^\mathrm{th}$ object, and let $\mathbf{f}_{t, 1:M}$ be the set of such features for all objects.
Since different objects can interact with each other, it is necessary to use a method that can inform each object about the effects of their interactions with other objects.
Moreover, since features extracted from different objects comprise a set, this method should be permutation-equivariant,\ie the results should not depend on the order in which object features are processed.
Therefore, we use the multi-head self-attention block (\textsc{sab}, \citet{Lee2019settransformer}), which is able to account for higher-order interactions between set elements when computing their representations. 
Intuitively, in our case, \textsc{sab} allows any of the trackers to query other trackers about attributes of their respective objects,\eg distance between objects, their direction of movement, or their relation to the camera.
This is implemented as follows,
{\small
\begin{align}
Q &= W_q \mathbf{f}_{1:M} + b_q\,, \qquad K = W_k \mathbf{f}_{1:M} + b_k\,, \qquad V = W_v \mathbf{f}_{1:M} + b_v \label{eq:projection}\,,\\
&\hspace{1.5em} O_i = \operatorname{softmax}\left( Q_i K_i^T / \sqrt{d_q} \right) V_i\,, \qquad i=1,\dots,H\,, \label{eq:att}\\
&\hspace{7em} o_{1:M} = O = \operatorname{concat}(O_i,\dots,O_H)\,, \label{eq:multihead}
\end{align}}
where $o_m$ is the output of the relational reasoning module for object $m$. Time-step subscripts are dropped to decrease clutter.
In Eq. \ref{eq:projection}, each of the extracted features $\mathbf{f}_{t,m}$ is linearly projected into a triplet of key $\mathbf{k}_{t,m}$, query $\mathbf{q}_{t,m}$ and value $\mathbf{v}_{t,m}$ vectors. Together, they comprise $K, Q$ and $V$ matrices with $M$ rows and $d_q, d_q, d_v$ columns, respectively.
$K, Q$ and $V$ are then split up into multiple heads $H \in \mathbb{N}_+$, which allows to query different attributes by comparing and aggregating different projection of features.
Multiplying $Q_iK_i^T$ in Eq. \ref{eq:att} allows to compare every query vector $\mathbf{q}_{t,m,i}$ to all key vectors $\mathbf{k}_{t,1:M,i}$, where the value of the corresponding dot-products represents the degree of similarity.
Similarities are then normalised via a $\operatorname{softmax}$ operation and used to aggregate values $V$.
Finally, outputs of different attention heads are concatenated in Eq. \ref{eq:multihead}.
\textsc{Sab} produces $M$ output vectors, one for each input, which are then concatenated with corresponding inputs and fed into separate \gls{LSTM}s for further processing, as in \gls{HART}---see \Cref{fig:teaser}.

\Gls{MOHART} is trained fully end-to-end, contrary to other approaches. 
It maintains a hidden state, which can contain information about the object's motion. One benefit is that one can simply feed black frames into the model to predict future trajectories. Our experiments show that the model learns to fall back on the motion model captured by the \gls{LSTM} in this case.

%% file: 3_validation.tex
\section{Validation on Simulated Data} 
\label{sec:experiment_toy}

\label{app:toy_deterministic}
\begin{figure}%
	\centering
	\includegraphics[width=0.99\textwidth]{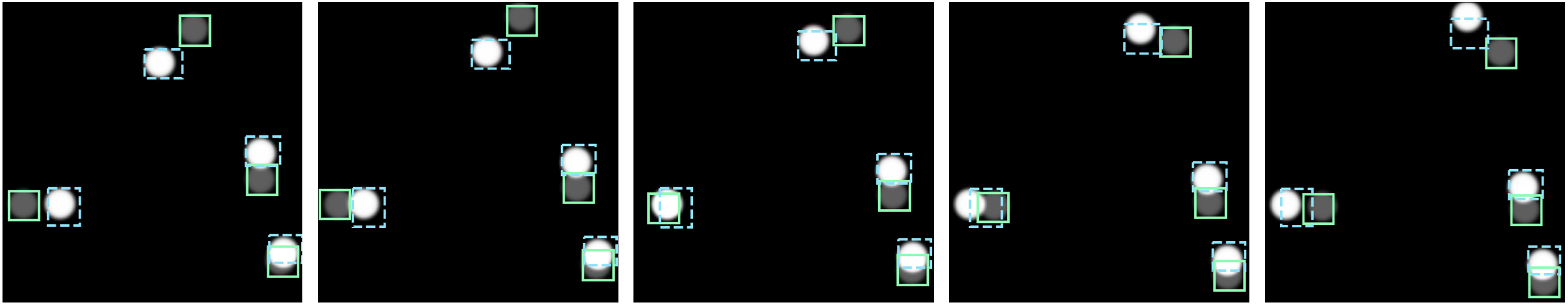}
	\caption{
		\textsc{hart} single object tracking applied four times in parallel and trained to predict the location of each circle three time steps into the future. Dashed lines indicate spatial attention, solid lines are predicted bounding boxes, faded circles show ground truth location at $T+3$. Each circle exerts repulsive forces on each other, where the force scales with $\nicefrac{1}{r}$, $r$ being their distance.
	}
	\label{fig:toy1}
\end{figure} 

We first evaluate the relational reasoning capabilities of the proposed algorithms on a toy domain---a two-dimensional square box filled with bouncing balls.
We train the model to predict future object locations (in contrast to simply tracking), to see how well the models understand motion patterns and interactions between objects. For details about the experimental setup, see \Cref{sec:appendix_toy}.

\subsection{First Experiment: Deterministic Domain}


In the first experiment in the toy domain (\Cref{fig:toy1}), four balls, which can be thought of as `protons in a box', repel each other.
\textsc{Hart} is applied four times in parallel and is trained to predict the location of each ball three time steps into the future.
Different forces from different objects lead to a non-trivial force field at each time step.
Accurately predicting the future location of an object using only its previous motion is therefore challenging (\Cref{fig:toy1} shows that each attention glimpse covers only the current object).
Surprisingly, the single object tracker solves this task with an average of $95\%$ IoU over sequences of 15 time steps, which shows the efficacy of end-to-end tracking to capture complex motion patterns and use them to predict future locations.
This, of course, could also be used to generate good-quality bounding boxes for a tracking task.

\subsection{Second Experiment: Stochastic Domain}
\label{app:toy_stochastic}

\begin{figure}[t!]%
	\centering
	\includegraphics[width=0.99\textwidth]{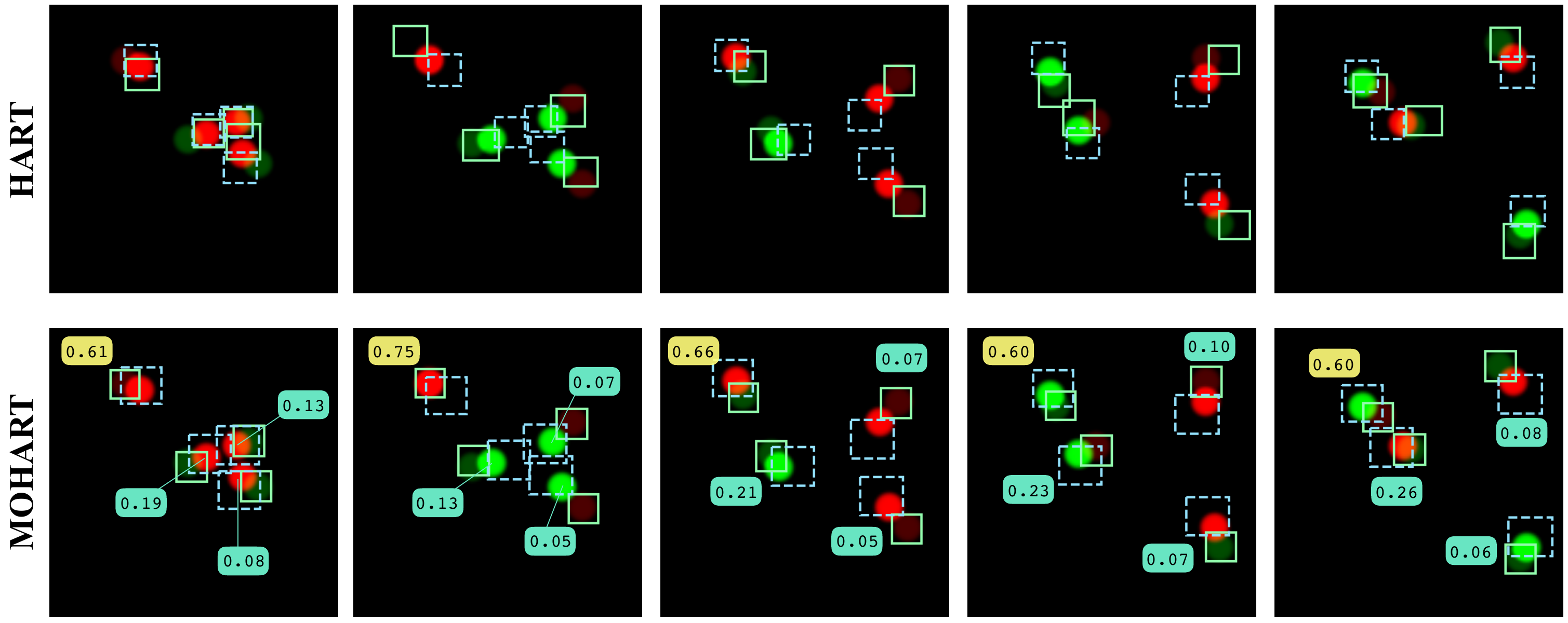}
	\caption{
		A scenario constructed to be impossible to solve without relational reasoning. Circles of the same colour repel each other, circles of different colour attract each other. Crucially, each circle is randomly assigned its identity in each time step. Hence, the algorithm can not infer the forces exerted on one object without knowledge of the state of the other objects in the current time step. The forces in this scenario scale with $1/\sqrt{r}$ and the algorithm was trained to predict one time step into the future. \textsc{hart} (top) is indeed unable to predict the future location of the objects accurately. The achieved average IoU is $47\%$, which is only slightly higher than predicting the objects to have the same position in the next time step as in the current one ($34\%$). Using the relational reasoning module, \textsc{mohart} (bottom) is able to make meaningful predictions ($76\%$ IoU). The numbers in the bottom row indicate the self-attention weights from the perspective of the top left tracker (yellow number box). Interestingly, the attention scores have a strong correlation with the interaction strength (which scales with distance) without receiving supervision.
		\vspace{-3mm}
	}
	\label{fig:toy2}
\end{figure}

\begin{figure}
    \centering
    \begin{subfigure}[c]{0.49\linewidth}
        \centering
        \includegraphics[width=\linewidth]{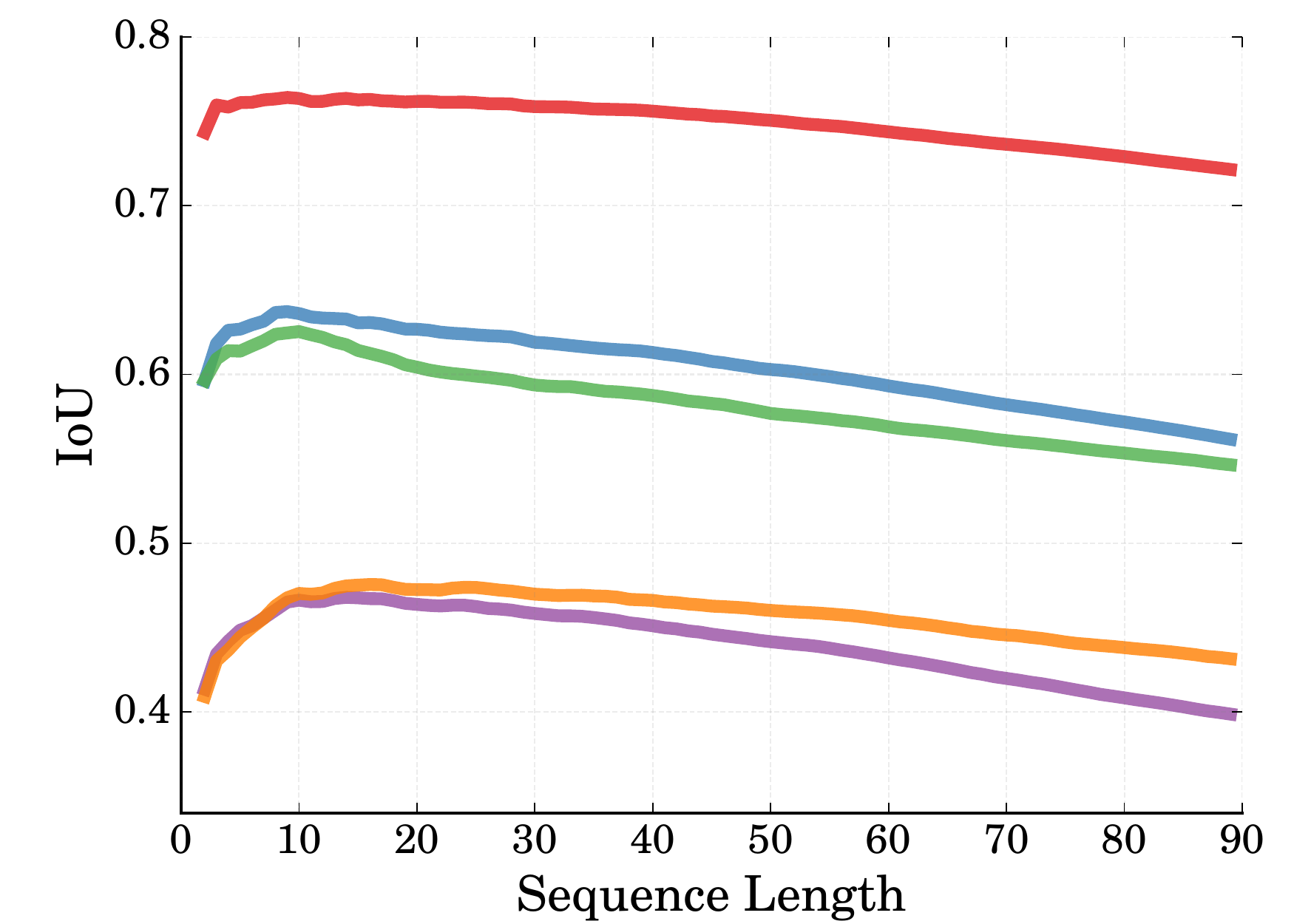}
    \end{subfigure}
    \begin{subfigure}[c]{0.49\linewidth}
        \centering
        \includegraphics[width=\linewidth]{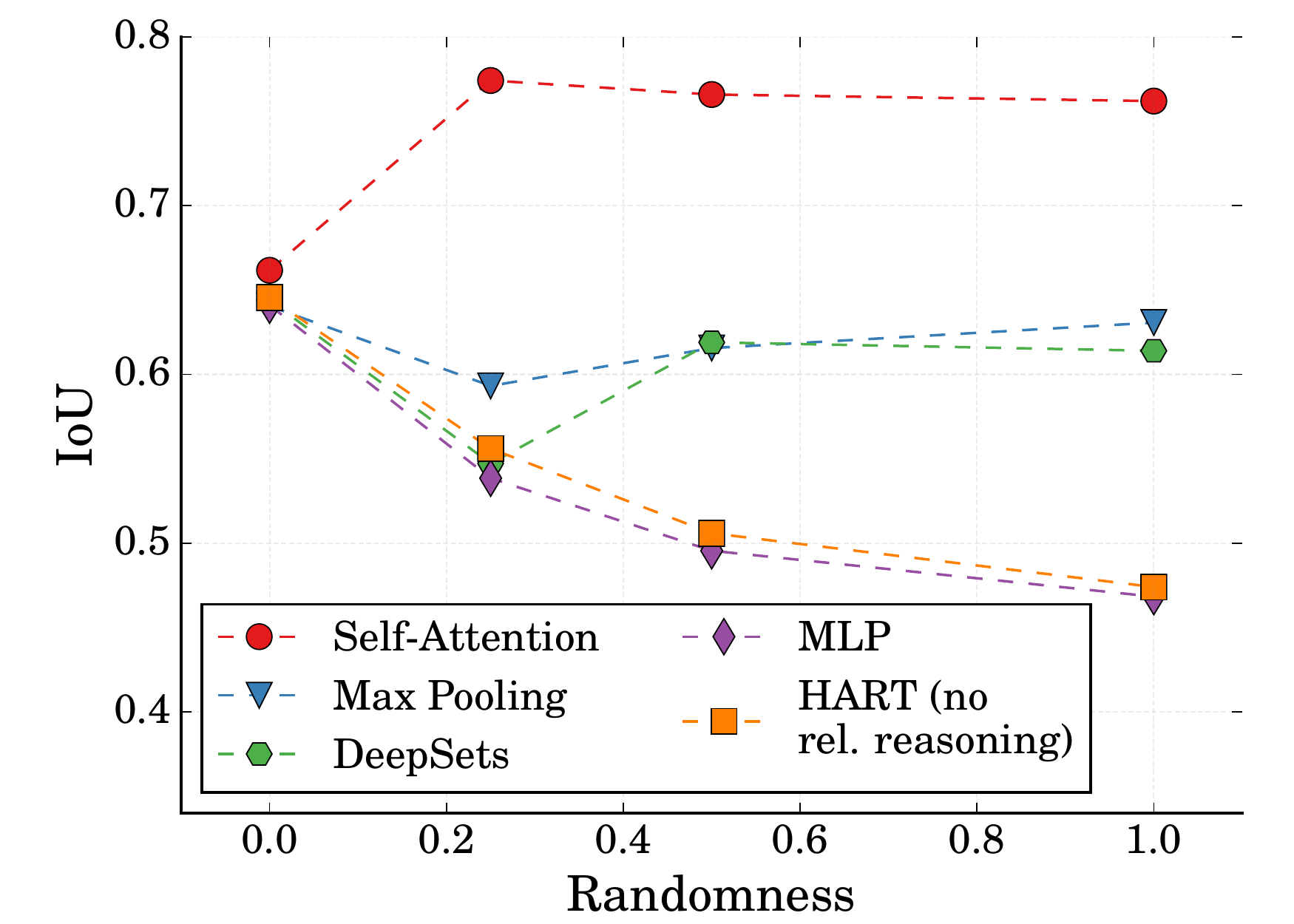}
    \end{subfigure}
    \vspace{-2mm}
    \caption{
        Left: average IoU over sequence length for different implementations of relational reasoning on the toy domain shown in \Cref{fig:toy2} ($\text{randomness} = 1.0$). Right: performance depends on \textit{randomness}---the frequency with which ball identities are randomly changed (sequence length 15).
        Higher randomness puts more pressure on relational reasoning.
        For $\text{randomness} = 0$, identities still have to be reassigned in some cases in order to prevent deadlocks, this leads to a performance loss for all models, which explains lower performance of self-attention for $\text{randomness} = 0$.
    }
    \label{fig:toy_quant}
\end{figure}




In the second experiment, we introduce randomness, rendering the scenario not solvable for a single object tracker as it requires knowledge about the state of the other objects and relational reasoning (see \Cref{fig:toy2}).
In each time step, we assign a colour-coded identity to the objects.
Objects of the same identity repel each other, object of different identities attract each other (the objects can be thought of as electrons and protons). We compare our proposed attention-based relational reasoning module to the following baselines:

\paragraph{\Gls{MLP}} In this version, the representations of all objects are concatenated and fed into a fully connected layer followed by ELU activations. The output is then again concatenated to the unaltered feature vector of each object. This concatenated version is then fed to the recurrent module of \gls{HART}. This way of exchanging information allows for universal function approximation (in the limit of infinite layer sizes) but does not impose permutation invariance.

\vspace{-1em}
\paragraph{DeepSets} Here, the learned representations of the different objects are summed up instead of concatenated and then divided by total number of objects. 
This is closely related to DeepSets \citep{Zaheer2017} and allows for universal function approximation of all permutation invariant functions \citep{Wagstaff2019}.

\vspace{-1em}
\paragraph{Max-Pooling} Similar to DeepSets, but using max-pooling as the permutation invariant operation. This way of exchanging information is used, e.g., by \citet{social-lstm} who predict future pedestrian trajectories from ground truth tracklets in coordinate space.

\Cref{fig:toy_quant} (left) shows a quantitative comparison of augmenting \textsc{hart} with different relational reasoning modules when identities are re-assigned in every timestep ($\text{randomness} = 1.0$).
Exchanging information between different trackers with an MLP leads to slightly worse performance than the baseline, while simple max-pooling performs significantly better ($\Delta \text{IoU} \sim 17\%$).
This can be explained through the permutation invariance of the problem: latent representation of different objects have no meaningful order; therefore the output of the model should be invariant to the ordering of the objects.
The MLP is in itself not permutation invariant and therefore prone to overfitting to the (meaningless) order of the objects in the training data.
Max-pooling, however, is permutation invariant and can in theory, despite its simplicity, be used to approximate any permutation invariant function given a sufficiently large latent space \citep{Wagstaff2019}.
Max-pooling is often used to exchange information between different tracklets,\eg in the trajectory prediction domain \citep{social-lstm,Gupta2019}.
However, self-attention, allowing for learned querying and encoding of information, solves the relational reasoning task much more accurately.

In \Cref{fig:toy_quant} (right), the frequency with which object identities are reassigned randomly is varied. The results show that, in a deterministic environment, tracking does not necessarily profit from relational reasoning - even in the presence of long-range interactions. The less random, the more static the force field is and a static force field can be inferred from a small number of observations (see \Cref{fig:toy1}). This does of course not mean that all stochastic environments profit from relational reasoning. What these experiments indicate is that tracking can not be expected to profit from relational reasoning by default in any environment, but instead in environments which feature (potentially non-deterministic) dynamics and predictable interactions.

%% file: 4_experiments.tex
\section{Relational Reasoning in Real-World Tracking}
\label{sec:experiment_real}

\begin{figure}[t!]
	\centering
	\includegraphics[width=1.0\textwidth]{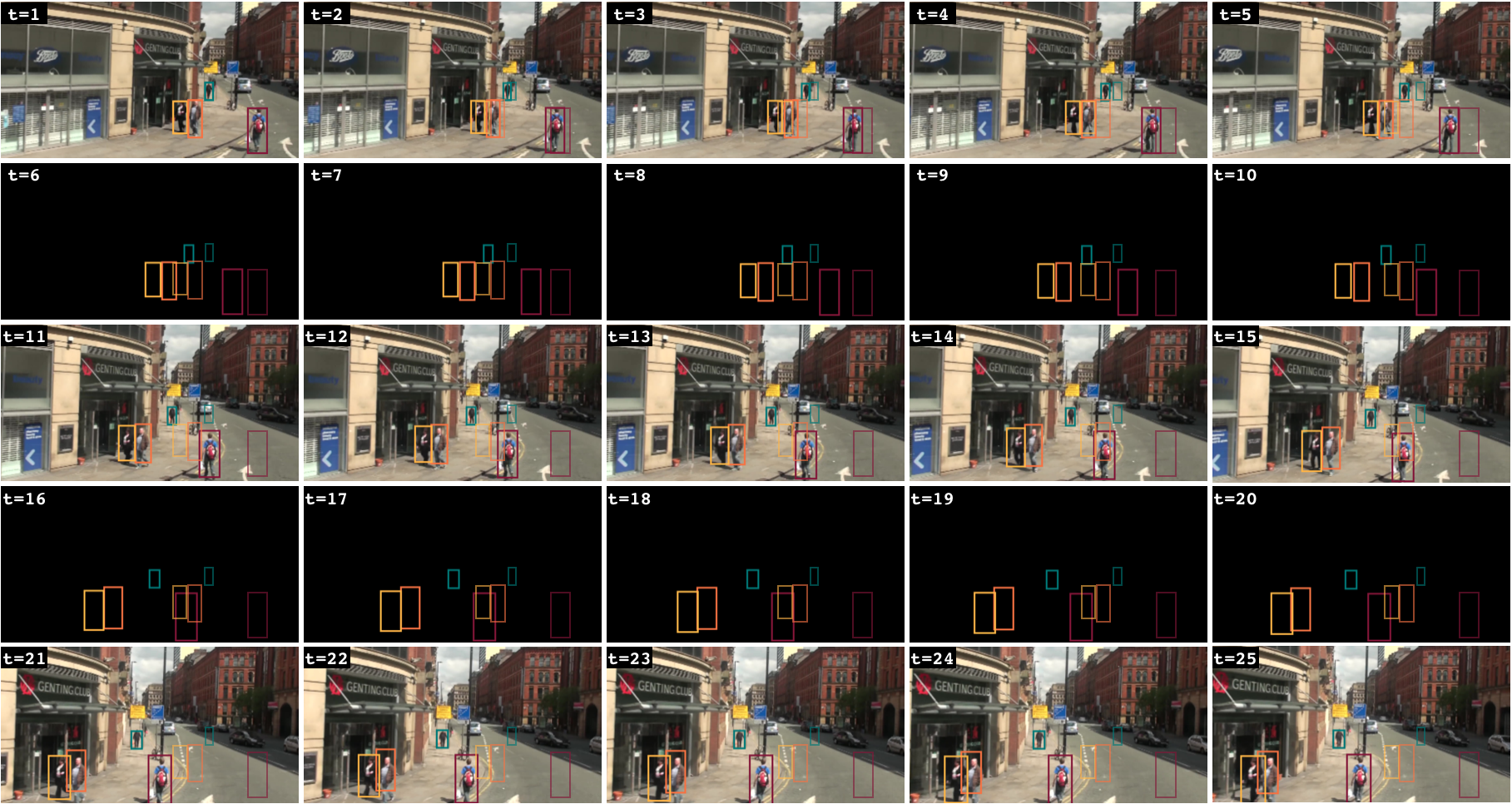}
	\vspace{-6mm}
	\caption{Camera blackout experiment on a street scene from the MOTChallenge dataset with strong ego-motion. Solid boxes are \textsc{mohart} predictions (for $t\geq 2$), faded bounding boxes indicate object locations in the first frame. As the model is trained end-to-end, \textsc{mohart} learns to fall back onto its internal motion model if no new observations are available (black frames). As soon as new observations come in, the model 'snaps' back onto the tracked objects.}
	\label{fig:blackout_main}
\end{figure}

Having established that \textsc{mohart} is capable of performing complex relational reasoning, we now test the algorithm on three real-world datasets and analyse the effects of relational reasoning on performance depending on dataset and task. We find consistent improvements of \textsc{mohart} compared to \textsc{hart} throughout. Relational reasoning yields particularly high gains for scenes with ego-motion, crowded scenes, and simulated faulty sensor inputs.

\subsection{Experimental Details}
\label{sec:exp_details}

We investigate three qualitatively  different datasets: the MOTChallenge dataset \citep{MOT16}, the UA-DETRAC dataset \citep{Wen15}, and the Stanford Drone dataset~\citep{DroneDataset}.
To increase scene dynamics and make the tracking/prediction problems more challenging, we sub-sample some of the high framerate scenes with a stride of two, resulting in scenes with 7-15 frames per second. Training and architecture details are given in \Cref{sec:experimental_details,sec:architecture_details}.
We conduct experiments in three different modes:

\textbf{Tracking.} The model is initialised with the ground truth bounding boxes for a set of objects in the first frame. It then consecutively sees the following frames and predicts the bounding boxes. The sequence length is 30 time steps and the performance is measured as \glsreset{IOU}\gls{IOU} averaged over the entire sequence excluding the first frame. This algorithm is either applied to the entire dataset or subsets of it to study the influence of certain properties of the data.

\textbf{Camera Blackout.} This simulates unsteady or faulty sensor inputs. The setup is the same as in 
\textit{Tracking}, but sub-sequences of the input are replaced with black images. The algorithm is expected to recognise that no new information is available and that it should resort to its internal motion model.

\textbf{Prediction.} Testing \textsc{mohart}'s ability to capture motion patterns, only two frames are shown to the model followed by three black frames. IoU is measured separately for each time step.

\begin{table}[!ht]

{\footnotesize

  \floatsetup{floatrowsep=qquad, captionskip=4pt}
        \ttabbox%
    {\begin{tabularx}{\textwidth}{l*{5}{>{\raggedleft\arraybackslash}X}}
      \toprule
&                   Entire &            Only &              No  &               Crowded &           Camera \\
&                   Dataset &           Ego-Motion &        Ego-Motion &        Scenes &            Blackout \\
      \midrule
\textbf{\gls{MOHART}} &   \textbf{68.5\%} &   \textbf{66.9\%} &   \textbf{64.7\%} &   \textbf{69.1\%} &   \textbf{63.6\%}  \\[0.1em]
\gls{HART} &              66.6\% &            64.0\% &            62.9\% &            66.9\% &            60.6\%  \\
\midrule
$\Delta$ &          1.9\% &             2.9\% &             1.8\% &             2.2\% &             3.0\%  \\ 
\bottomrule
      \addlinespace
      \addlinespace
      \addlinespace
      \end{tabularx}}
    {\caption{Tracking performance on the MOTChallenge dataset measured in IoU.}
      \label{tab:results_motc}}
  \begin{floatrow}[2]

    \ttabbox%
    {\begin{tabularx}{0.55\textwidth}{l*{3}{>{\raggedleft\arraybackslash}X}}
      \toprule
      & All & Crowded Scenes & Camera Blackout \\
      \midrule
      \textbf{\gls{MOHART}} & 68.1\% & \textbf{69.5\%} & \textbf{64.2\%}\\
      [0.1em]
      \gls{HART} & \textbf{68.4\%} & 68.6\% & 53.8\%\\
      \midrule
      $\Delta$ & -0.3\% & 0.9\% & 0.4\%\\
      \bottomrule
      \end{tabularx}}
    {\caption{UA-DETRAC Dataset}
      \label{tab:results_detrac}}
    \hfill%
    \ttabbox%
    {\begin{tabularx}{0.35\textwidth}{r*{3}{>{\raggedleft\arraybackslash}X}}
      \toprule
      All & Camera Blackout & CamBlack Bikes \\
      \midrule
      \textbf{57.3\%} & \textbf{53.3}\% & \textbf{53.3\%} \\
      [0.1em]
      56.1\% & 52.6\% & 50.7\%\\
      \midrule
      1.2\% & 0.7\% & 2.6\%\\
      \bottomrule
      \end{tabularx}}
    {\caption{Stanford Drone Data}
      \label{tab:results_stanford}}
  \end{floatrow}
  
  }
\end{table}%

\subsection{Results and Analysis}

On the MOTChallenge dataset, \textsc{hart} achieves $66.6\%$ \gls{IOU} (see \Cref{tab:results_motc}), which in itself is impressive given the small amount of training data of only 5225 training frames and no pre-training. \textsc{mohart} achieves $68.5\%$ (both numbers are averaged over 5 runs, independent samples $t$-test resulted in $p < 0.0001$). The performance gain increases when only considering ego-motion data. This is readily explained: movements of objects in the image space due to ego-motion are correlated and can therefore be better understood when combining information from movements of multiple objects, i.e. performing relational reasoning. In another ablation, we filtered for only crowded scenes by requesting five objects to be present for, on average, 90\% of the frames in a sub-sequence. For the MOT-Challenge dataset, this only leads to a minor increase of the performance gain of \textsc{mohart} indicating that the dataset exhibits a sufficient density of objects to learn interactions. The biggest benefit from relational reasoning can be observed in the \textit{camera blackout} experiments (setup explained in \Cref{sec:exp_details}). Both \textsc{hart} and \textsc{mohart} learn to rely on their internal motion models when confronted with black frames and propagate the bounding boxes according to the previous movement of the objects. It is unsurprising that this scenario profits particularly from relational reasoning. Qualitative tracking and \textit{camera blackout} results are shown in \Cref{fig:blackout_main} and \Cref{sec:blackout}.

Tracking performance on the UA-DETRAC dataset only profits from relational reasoning when filtering for crowded scenes (see \Cref{tab:results_detrac}). The fact that the performance of \textsc{mohart} is slightly worse on the vanilla dataset ($\Delta = -0.3\%$) can be explained with more overfitting. As there is no exchange between trackers for each object, each object constitutes an independent training sample.

The Stanford drone dataset (see \Cref{tab:results_stanford}) is different to the other two---it is filmed from a birds-eye view.
The scenes are more crowded and each object covers a small number of pixels, rendering it a difficult problem for tracking.
The dataset was designed for trajectory prediction---a setup where an algorithm is typically provided with ground-truth tracklets in coordinate space and potentially an image as context information. The task is then to extrapolate these tracklets into the future. 
The performance gain of \textsc{mohart} on the \textit{camera blackout} experiments is particularly strong when only considering cyclists. 

In the \textit{prediction} experiments (see \Cref{sec:exp_pred}), \textsc{mohart} consistently outperforms \textsc{hart}. On both datasets, the model outperforms a baseline which uses momentum to linearly extrapolate the bounding boxes from the first two frames. This shows that even from just two frames, the model learns to capture motion models which are more complex than what could be observed from just the bounding boxes (i.e. momentum), suggesting that it uses visual information (\textsc{hart} \& \textsc{mohart}) as well as relational reasoning (\textsc{mohart}).

\subsection{Visual Object Tracking Performance}
\label{sec:sota}

We compare our method to \gls{SOTA} visual object trackers on the MOTChallege dataset.
While the evaluation is identical for all methods (see \Cref{sec:experimental_details}), the training data differs.
\Gls{MOHART} was trained on the MOTChallenge dataset (excluding the evaluation frames), while the other methods were trained on different datasets\footnote{as specified in Tab.\ 10 of \citet{SiamRCNN}; up to $2.3$ million frames compared to $\approx$8k used for \gls{MOHART}}, with three orders of magnitude more data in total, but they were \textit{not} trained on any part of the MOTChallenge.
Fine-tuning the baselines on the target dataset is therefore likely to improve their performance.
\Cref{tab:sota} shows that \gls{MOHART} performs worse but comparable to previous \gls{SOTA} models \cite{SiamRPNpp, ECO, DIMP}.
It is, however, significantly outperformed by the concurrently developed \gls{SIAMRCNN} \cite{SiamRCNN}.

\begin{table}[!ht]
	
	{\footnotesize
		
		\floatsetup{floatrowsep=quad, captionskip=4pt}
		\ttabbox%
		{\begin{tabularx}{\textwidth}{l*{5}{>{\raggedleft\arraybackslash}X}}
				\toprule
				& \gls{MOHART} &  \gls{SIAMRPN}\cite{SiamRPNpp} & \gls{ECO} \cite{ECO}& \gls{DIMP}-50 \cite{DIMP} & \gls{SIAMRCNN} \cite{SiamRCNN}\\
				\midrule
				\textsc{iou} &   72.1\% &   75.8\% &   74.6\% &  76.2\% & \textbf{85.1\%}  \\[0.1em]
		\end{tabularx}}
		{\caption{Comparison to non-end-to-end$^3$ \gls{SOTA} algorithms for \gls{VOT} on the MOTChallenge dataset. Here, the original framerate was used leading to higher than performance in \Cref{tab:results_motc}.}			\label{tab:sota}}
	}
	\vspace{-1em}
\end{table}%

%% file: 5_discussion.tex
\vspace*{-1em}
\section{Conclusion}
\label{sec:discussion}

With \gls{MOHART}, we introduce an end-to-end multi-object tracker that is capable of capturing complex interactions and leveraging these for precise predictions as experiments both on toy and real-world data show. However, the experiments also show that the benefit of relational reasoning strongly depends on the nature of the data. 
In particular, the toy experiments showed that in an entirely deterministic world, relational reasoning was much less important than in a stochastic environment.
Amongst the real-world dataset, the highest performance gains from relational reasoning were achieved on the MOTChallenge dataset, which features crowded scenes, ego-motion and occlusions.
Compared to \gls{SOTA} visual object trackers, \gls{MOHART} achieves inferior-but-comparable performance.
The relational reasoning toy experiments, the camera blackout and the prediction experiments, as well as the training on a fraction of the data, show the flexibility and the potential of our approach compared to the non-end-to-end\footnote{end-to-end meaning gradients are propagated through all parts of the model and across the time axis} \gls{SOTA} visual object trackers.
We see two potential routes for further improving the performance of \gls{MOHART} in the future: (1) leveraging pre-training on other datasets---this point might seem trivial, but our initial experiments demonstrated no performance improvements---and (2) incorporating external detections, which would also allow for a fair comparison against tracking-by-detection methods.

%% file: 6_experimental_details.tex
\section{Experimental Details}
\label{sec:experimental_details}
The MOTChallenge and the UA-DETRAC dataset discussed in this section are intended to be used as a benchmark suite for multi-object-tracking in a tracking-by-detection paradigm. Therefore, ground truth bounding boxes are only available for the training datasets. The user is encouraged to upload their model which performs tracking in a data association paradigm leveraging the provided bounding box proposals from an external object detector. As we are interested in a different analysis (IoU given inital bounding boxes), we divide the training data further into training and test sequences. To make up for the smaller training data, we extend the MOTChallenge 2017 dataset with three sequences from the 2015 dataset (ETH-Sunnyday, PETS09-S2L1, ETH-Bahnhof). We use the first 70\% of the frames of each of the ten sequences for training and the rest for testing. Sequences with high frame rates (30Hz) are sub-sampled with a stride of two. For the UA-DETRAC dataset, we split the 60 available sequences into 44 training sequences and 16 test sequences. For the considerably larger Stanford Drone dataset we took three videos of the scene \textit{deathCircle} for training and the remaining two videos from the same scene for testing. The videos of the drone dataset were also sub-sampled with a stride of two to increase scene dynamics.

For \Cref{sec:sota}, all methods were evaluated on the first 30 frames of sequences '02', '05', '09', '11' of the MOTChallenge dataset using all objects visible in the first frame. We used the original framerate of the dataset. For details about architecture and training set-up of the baselines, please see \citet{SiamRCNN}.

%% file: 6_architecture_details.tex
\section{Architecture Details}
\label{sec:architecture_details}
The architecture details were chosen to optimise \textsc{hart} performance on the MOTChallenge dataset. They deviate from the original \textsc{hart} implementation \citep{Kosiorek17} as follows: A presence variable predicts whether an object is in the scene and successfully tracked. This is trained with a binary cross entropy loss. The maximum number of objects to be tracked simultaneously was set to 5 for the UA-DETRAC and MOTChallenge dataset. For the more crowded Stanford drone dataset, this number was set to 10. The feature extractor is a three layer convolutional network with a kernel size of 5, a stride of 2 in the first and last layer, 32 channels in the first two layers, 64 channels in the last layer, ELU activations, and skip connections. This converts the initial $32 \times 32 \times 3$ glimpse into a $7 \times 7 \times 64$ feature representation. This is followed by a fully connected layer with a 128 dimensional output and an elu activation. The spatial attention parameters are linearly projected onto 128 dimensions and added to this feature representation serving as a positional encoding. The LSTM has a hidden state size of 128. The self-attention unit in \textsc{mohart} comprises linear projects the inputs to dimensionality 128 for each keys, queries and values. For the real-world experiments, in addition to the extracted features from the glimpse, the hidden states from the previous LSTM state are also fed as an input by concatinating them with the features. In all cases, the output of the attention module is concatenated to the input features of the respective object.

As an optimizer, we used RMSProp with momentum set to $0.9$ and learning rate $5*10^{-6}$. For the MOTChallenge dataset and the UA-DETRAC dataset, the models were trained for 100,000 iterations of batch size 10 and the reported IoU is exponentially smoothed over iterations to achieve lower variance. For the Stanford Drone dataset, the batch size was increased to 32, reducing time to convergence and hence model training to 50,000 iterations.

%% file: 6_first_toy_experiment.tex
\section{Bouncing Balls}
\label{sec:appendix_toy}

The toy domain consists of a square two-dimensional box filled with bouncing balls.
The balls move and can collide with each other with approximated elastic collisions (energy and momentum conservation).
Additionally, balls exert either repulsive force (first experiment) or repulsive/attractive force (second experiment, colour-coded), which scales with $1/r$, $r$ being the distance between centres of the balls.



%% file: 6_exp_prediction.tex
\section{Prediction Experiments}
\label{sec:exp_pred}

\begin{figure}
    \centering
    \begin{subfigure}[c]{0.99\linewidth}
        \centering
        \includegraphics[width=\linewidth]{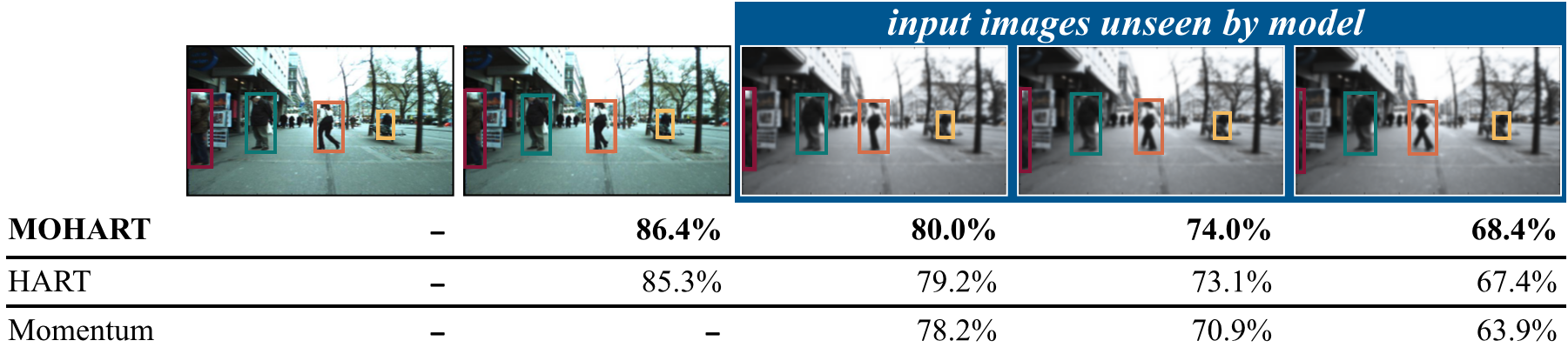}
        \vspace{-6mm}
        \caption{Prediction results on the MOTChallenge dataset \cite{MOT16}.}
        \label{fig:MOTC_imgs}
    \end{subfigure}
    \vspace{2mm}
    \begin{subfigure}[c]{0.99\linewidth}
        \centering
        \includegraphics[width=\linewidth]{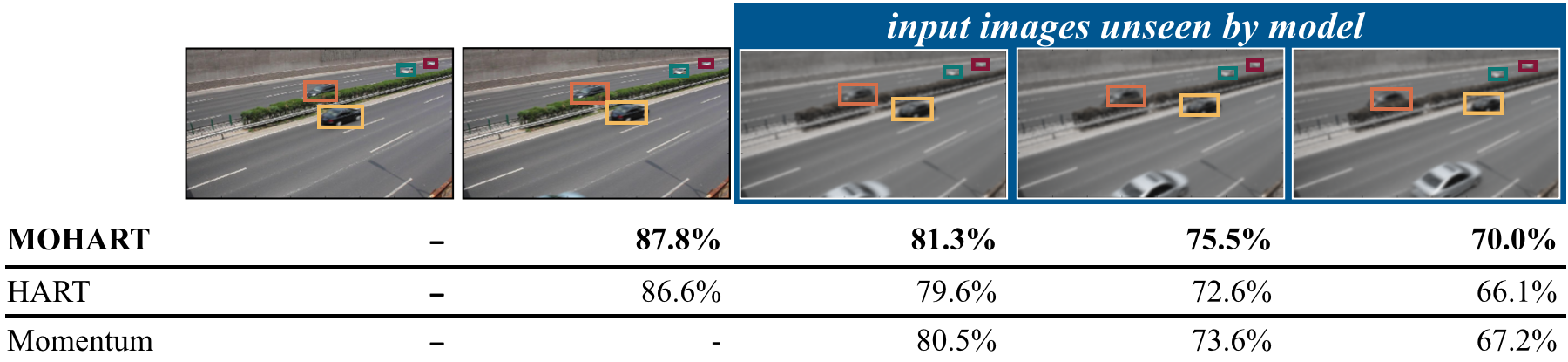}
        \vspace{-6mm}
        \caption{Prediction results on the UA-DETRAC dataset (crowded scenes only) \cite{Wen15}.}
        \label{fig:Detrac_quant}
    \vspace{-5mm}
    \end{subfigure}
\caption{Peeking into the future. Only the first two frames are shown to the tracking algorithm followed by three black frames. \textsc{mohart} learns to fall back on its internal motion model when no observation (i.e. only a black frame) is available. The reported IoU scores show the performance for the respective frames 0, 1, 2, and 3 time steps into the future.
\label{fig:prediction}
}
\end{figure}

In the results from the \textit{prediction} experiments (see \Cref{fig:prediction}) \textsc{mohart} consistently outperforms \textsc{hart}. On both datasets, the model outperforms a baseline which uses momentum to linearly extrapolate the bounding boxes from the first two frames. This shows that even from just two frames, the model learns to capture motion models which are more complex than what could be observed from just the bounding boxes (i.e. momentum), suggesting that it uses visual information (\textsc{hart} \& \textsc{mohart}) as well as relational reasoning (\textsc{mohart}). The strong performance gain of \textsc{mohart} compared to \textsc{hart} on the UA-DETRAC dataset, despite the small differences for tracking on this dataset, can be explained as follows: this dataset features little interactions but strong correlations in motion. Hence when only having access to the first two frames, \textsc{mohart} profits from estimating the velocities of multiple cars simultaneously.

%% file: 6_blackout.tex
\section{Qualitative Tracking Results}
\label{sec:blackout}

\begin{figure}[ht!]
\centering
\includegraphics[width=1\textwidth]{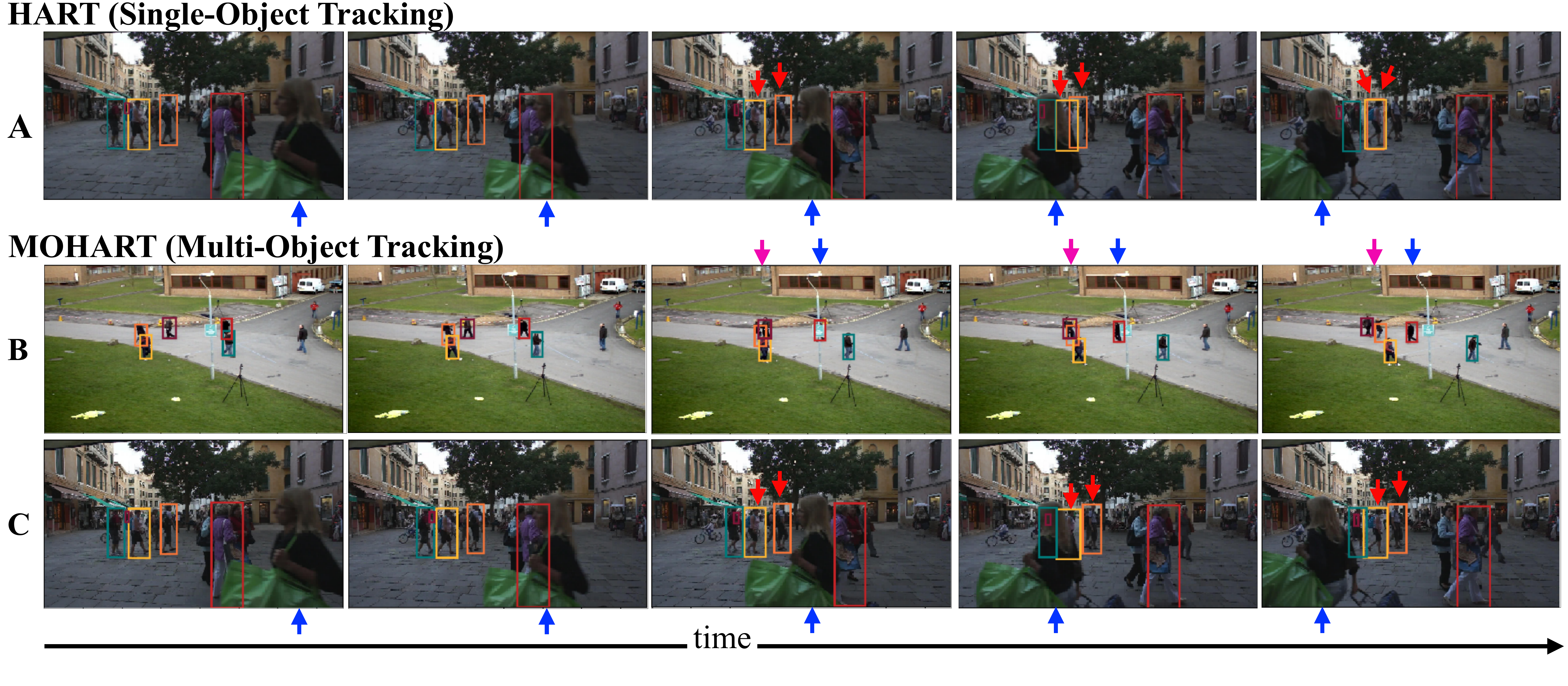}
\vspace{-8mm}
\caption{Tracking examples of both \textsc{hart} and \textsc{mohart}. Coloured boxes are bounding boxes predicted by the model, arrows point at challenging aspects of the scenes. (A) \& (C): Each person being tracked is temporarily occluded by a woman walking across the scene (blue arrows). \textsc{mohart}, which includes a relational reasoning module, handles this more robustly (compare red arrows).
\vspace{-4mm}}
\label{fig:tracking_and_predicting}
\end{figure}

\appendix
\begin{figure}
	\centering
	\includegraphics[width=1.0\textwidth]{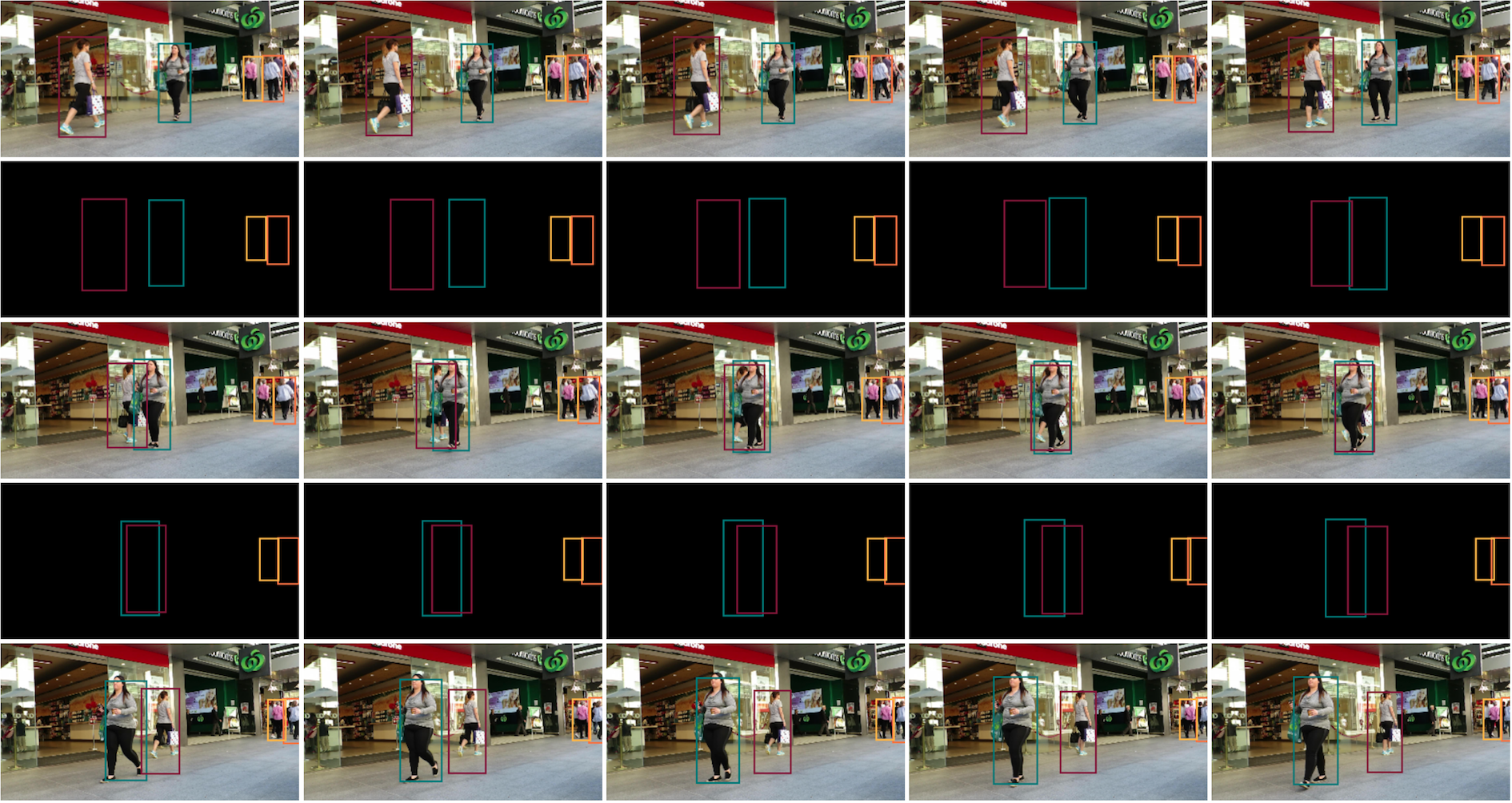}
	\vspace{-6mm}
	\caption{Camera blackout experiment on a pedestrian street scene from the MOTChallenge dataset without ego-motion. Subsequent frames are displayed going from top left to bottom right. Shown are the inputs to the model (some of them being black frames, i.e. arrays of zeroes) and bounding boxes predicted by \textsc{MOHART} (coloured boxes). This scene is particularly challenging as occlusion and missing sensor input coincide (fourth row).
		\vspace{-2mm}}
	\label{fig:blackout1}
\end{figure}


In \Cref{sec:experiment_real}, we tested \textsc{mohart} on three different real world data sets and in a number of different setups. \Cref{fig:tracking_and_predicting} shows qualitative results both for \textsc{hart} and \textsc{mohart} on the MOTChallenge dataset.

Furthermore, we conducted a set of camera blackout experiments to test \textsc{mohart}'s capability of dealing with faulty sensor inputs. While traditional pipeline methods require careful consideration of different types of corner cases to properly handle erroneous sensor inputs, \textsc{mohart} is able to capture these automatically, especially when confronted with similar issues in the training scenarios. To simulate this, we replace subsequences of the images with black frames. \Cref{fig:blackout1} and \Cref{fig:blackout_main} show two such examples from the test data together with the model's prediction. \textsc{mohart} learns not to update its internal model when confronted with black frames and instead uses the LSTM to propagate the bounding boxes. When proper sensor input is available again, the model uses this to make a rapid adjustment to its predicted location and `snap' back onto the object. This works remarkably well in both the presence of occlusion (\Cref{fig:blackout1}) and ego-motion (\Cref{fig:blackout_main}). \Cref{tab:results_motc,tab:results_detrac,tab:results_stanford} show that the benefit of relational reasoning is particularly high in these scenarios specifically. These experiments can also be seen as a proof of concept of \textsc{mohart}'s capabalities of predicting future trajectories---and how this profits from relational reasoning.